\documentclass[conference]{IEEEtran}
\IEEEoverridecommandlockouts

\usepackage{cite}
\usepackage{amsmath,amssymb,amsfonts}
\usepackage{algorithmic}
\usepackage{graphicx}
\usepackage{textcomp}
\usepackage{xcolor}
\usepackage{epstopdf}
\usepackage{times}
\usepackage{soul}
\usepackage{url}
\usepackage[hidelinks]{hyperref}
\usepackage[utf8]{inputenc}
\usepackage[small]{caption}
\usepackage{graphicx}
\usepackage{amsmath}
\usepackage{booktabs}
\usepackage{algorithm}
\usepackage{algorithmic}
\usepackage{times}
\usepackage{epsfig}
\usepackage{graphicx}
\usepackage{amsmath}
\usepackage{amssymb}
\usepackage{booktabs}       
\usepackage{amsfonts}       
\usepackage{nicefrac}       
\usepackage{microtype}      
\usepackage{tabularx}
\usepackage{algorithm}
\usepackage{algorithmic}
\usepackage{epsfig}
\usepackage{amsmath}
\usepackage{multirow}
\usepackage{subfigure}

\def\BibTeX{{\rm B\kern-.05em{\sc i\kern-.025em b}\kern-.08em
    T\kern-.1667em\lower.7ex\hbox{E}\kern-.125emX}}
\begin{document}

\title{Semi-supervised Learning with Adaptive Neighborhood Graph Propagation Network}

\author{\IEEEauthorblockN{Bo Jiang, Leiling Wang, Jin Tang and Bin Luo}
\IEEEauthorblockA{\textit{School of Computer Science and Technology, Anhui University} \\
jiangbo@ahu.edu.cn}

}

\maketitle

\begin{abstract}

Graph methods have been commonly employed for visual data representation and analysis in computer vision area.
Graph  data representation and learning becomes more and more important in this area.
Recently, Graph Convolutional Networks (GCNs) have been employed for graph data representation and learning. 
Existing GCNs usually use a fixed neighborhood graph which is not guaranteed to be optimal for semi-supervised learning tasks.
In this paper, we  propose a unified adaptive neighborhood feature propagation model
and  derive a novel Adaptive Neighborhood Graph Propagation Network (ANGPN) for data representation and semi-supervised learning.
The aim of ANGPN is to
conduct both neighborhood graph construction (from pair-wise distance information) and graph convolution simultaneously in a unified formulation and
thus can learn an optimal neighborhood graph that best serves graph convolution for graph based semi-supervised learning.
Experimental results demonstrate the effectiveness and benefit of the proposed ANGPN on image data semi-supervised classification tasks.

\end{abstract}

\section{Introduction}

Compact data representation and learning is a fundamental problem in computer vision area.
Graph methods have been commonly employed for visual data representation and analysis, such as classification, segmentation, etc.
The core technique in graph methods is graph representation and learning.
Recently, there is an increasing attention on trying to generalize CNNs to Graph Convolutional Networks (GCNs) to deal with graph  data~\cite{defferrard2016convolutional}\cite{monti2017geometric} \cite{hamilton2017inductive} \cite{velickovic2017graph}.
For example,
Bruna et al.~\cite{bruna2014spectral} employ a eigen-decomposition of graph Laplacian matrix to define graph convolution operation. 
Henaff et al.~\cite{henaff2015deep} introduce a spatially constrained spectral filters to define graph convolution. 
Defferrard et al.~\cite{defferrard2016convolutional} propose to utilize Chebyshev expansion to approximate the spectral filters for graph convolution operation. 
Kipf et al.~\cite{kipf2016semi}  present a simple Graph Convolutional Network (GCN). 
The main idea of GCN is to explore the first-order approximation of spectral filters to compute graph convolution more efficiently.
Monti et al.~\cite{monti2017geometric} propose mixture model CNNs (MoNet) for graph data learning and analysis. 
Veli{\v{c}}kovi{\'c} et al.~\cite{velickovic2017graph} propose Graph Attention Networks (GAT) for graph based semi-supervised learning which can aggregate the features of neighboring nodes in an adaptive weighting manner.
Hamilton et al.~\cite{hamilton2017inductive} present  a general inductive representation and learning framework (GraphSAGE) by sampling and
 aggregating features from a node’s local neighborhood.
Klicpera et al.~\cite{klicpera2018predict} propose to  integrate PageRank propagation into GCN in layer-wise propagation.
Petar et al.~\cite{velickovic2018deep} propose Deep Graph Infomax (DGI) to learn the node representation in unsupervised manner.
Jiang et al.,~\cite{jiang2019data} recently propose Graph Diffusion-Embedding Network (GEDNs) for graph node semi-supervised learning. 

The above state-of-the-art methods can be widely used in many graph learning tasks, such as graph based semi-supervised learning, low-dimensional representation, clustering etc. 
In computer vision applications, for these learning tasks, existing  methods usually need to use a two-stage framework, i.e., 1) obtaining/constructing a neighborhood graph from data and
 2) conducting graph convolutional representation and learning on this graph.
However, one main issue is that such a two-stage framework does not
fully mine the correlation between \emph{graph construction} and  \emph{graph convolutional learning}, which may lead to
weak suboptimal solution.
Previous works generally focus on graph convolution operation while little attention has been put on graph construction.
To address this issue, Li et al.~\cite{adaptive_GCN} present an optimal graph CNNs, in which the graph is learned adaptively by employing a metric learning method.
Veli{\v{c}}kovi{\'c} et al.~\cite{velickovic2017graph} propose Graph Attention Networks (GAT) for graph based semi-supervised learning.
Jiang et al.~\cite{jiang2019graph} propose Graph Optimized Convolutional Network (GOCN) by integrating graph construction and convolution together in a unified operation model. 
They also~\cite{jiang2019mask} propose Graph mask Convolutional Network (GmCN) by selecting neighbors adaptively in GCN operation. 


In this paper, motivated by recent graph learning~\cite{Nie2014Clustering,nie2016unsupervised} and graph optimized GCN works~\cite{jiang2019mask,jiang2019graph,klicpera2018predict},  we propose a novel graph neural network, named Adaptive Neighborhood  Graph Propagation Network (ANGPN), for graph based semi-supervised learning task.
Similar to~\cite{jiang2019graph,jiang2019mask}, ANGPN aims to conduct both  neighborhood graph learning and  graph convolution together and formulate  graph construction and convolution simultaneously.
Different from previous works~\cite{jiang2019graph,jiang2019mask,velickovic2017graph}, ANGPN  aims to learn an optimal neighborhood graph for graph convolution from pair-wise distance information and can be used in many computer vision tasks which do not have any prior graph structures.
Overall, the main contributions of this paper are summarized as follows:
\begin{itemize}
  \item  We propose a novel Adaptive Neighborhood Graph Propagation Network (ANGPN) which integrates both neighborhood graph construction and graph  convolution cooperatively in a unified formulation for data representation and semi-supervised learning problem.
  \item  We propose a novel model of Adaptive Neighborhood Feature Propagation (ANFP)  which provides a context-aware  representation for data by jointly employing both unary feature of each data and contextual features from its neighbors together.
\end{itemize}
Experimental results on several datasets demonstrate the effectiveness of ANGPN on semi-supervised learning tasks. 

\section{Adaptive Neighborhood Feature Propagation}

In the following, we first introduce a neighborhood graph feature  propagation (NFP) model. Then
we extend it to Adaptive NFP (ANFP) by constructing the neighborhood graph in NFP adaptively.

\subsection{Neighborhood feature propagation}

Let $M\in \mathbb{R}^{n\times n}$ be the adjacency matrix of a neighborhood graph with $M_{ii}=0$, i.e.,
if node $v_j\in \mathcal{N}(v_i)$ then $M_{ij}=1$, otherwise $M_{ij}=0$.
Let ${A}=MD^{-1}$ be the row-normalization of $M$, where $D$ is a diagonal matrix with $D_{ii}=\sum_j M_{ij}$.
Thus, we have $\sum^n_{j=1}A_{ij}=1, A_{ij}\geq 0$ and $A_{ii}=0$.
Let  ${H}=({h}_1,{h}_2 \cdots {h}_n)\in \mathbb{R}^{n\times d}$ be the feature descriptors of graph nodes.
Then, the aim of our \emph{Neighborhood Graph based Feature Propagation} is to
learn a kind of feature representation ${F}=\mathcal{P}({A}, {H})=({f}_1, {f}_2\cdots {f}_n)\in \mathbb{R}^{n\times d}$
for graph nodes by incorporating the contextual information of their neighbors.
Here, we propose a scheme that is similar to the one in label propagation~\cite{wang2008label} but iteratively propagates the \textbf{features} $H$ on neighborhood graph $A$.
Formally, for each node $v_i$, we absorb a fraction of feature information from its neighbors  in
each  propagation step as follows,
\begin{equation}\label{EQ:propagation_rwr00}
{f}_i^{(t+1)} = \alpha \sum^n\nolimits_{j=1}{A}_{ij} {f}_j^{(t)} + (1-\alpha){h}_i
\end{equation}
where $t=0, 1\cdots$ and $f^{(0)}_j = h_j$. Parameter $\alpha \in (0, 1)$ is the fraction of feature information that node $v_i$ receives from
its neighbors.
%
We can rewrite Eq.(7) more compactly as
\begin{equation}\label{EQ:propagation_rwr}
{F}^{(t+1)} = \alpha {A} {F}^{(t)} + (1-\alpha){H}
\end{equation}
where ${F}^{(0)}=H$.
It is known that the above propagation will converge to an equilibrium solution as
\begin{equation}\label{EQ:propagation_rwr0}
{F}^{*} = (1-\alpha) (I - \alpha A)^{-1}{H}
\end{equation}

\subsection{Adaptive neighborhood feature propagation}

In the above feature propagation, one important aspect is the construction of neighborhood graph $A$.
One popular way is to construct a human established graph, such as k-nearest neighbor graph. 
However, such a pre-defined graph may have no clear structure and thus also be not guaranteed to best serve the feature propagation and learning task. 
To overcome this issue, we propose to learn an adaptive neighborhood graph ${S}$ to better capture the intrinsic neighborhood relationship among data in the above NFP. We call it as Adaptive Neighborhood Feature Propagation (ANFP).

In order to do so,
we first show that the above propagation
has an equivalent  optimization formulation~\cite{jiang2019data,wang2008label,jiang2019graph}. 
In particular, the converged solution of Eq.(\ref{EQ:propagation_rwr0}) is the optimal solution
that minimizes the following optimization problem,
\begin{equation}\label{EQ:propagation_opt}
\min_{F} \,\, \mathrm{Tr} (F^{T}(I - A) F) + \mu \|F - H\|^2_F
\end{equation}
where $\alpha = \frac{1}{1+\mu}$. $\mathrm{Tr}(\cdot)$ denotes the trace function and $\|\cdot\|_F$ denotes the Frobenious norm of matrix.
Then, inspired by~\cite{Nie2014Clustering,nie2016unsupervised}, we incorporate graph learning into Eq.(4) and  construct a unified model as 
\begin{align}\label{EQ:propagation_opt}
&\min_{S,F}\,\, \sum^n\nolimits_{i,j=1}D^x_{ij}S_{ij}+\gamma \|S\|^2_F+ \beta \mathrm{Tr} (F^{T}(I - S) F)\nonumber \\
&\,\,\,\,\,\,\,\,\,\,\,\, + \mu \|F - H\|^2_F\\
& s.t. \ \ \sum^n\nolimits_{j=1}S_{ij}=1, S_{ij}\geq 0
\end{align}
where $D^x_{ij}=\|x_i-x_j\|_2$ denotes the Euclidean distance between original input feature $x_i$ and $x_j$ of data\footnote{For efficiency consideration, we use the original feature $X$ to compute $D^x$ which is fixed in the following training process. Here, one can also use the current learned representation $H$ to obtain/update more accurate $D^h$.}. 
The optimal $S_{ij}$ can be regarded as a confidence (or probability) that
data $x_j$ is connected to $x_i$ as a neighbor.
A larger distance $D^x_{ij}$ should be assigned with smaller confidence $S_{ij}$.
The parameter $\gamma > 0$ is used to control the sparsity of learned graph $S$~\cite{Nie2014Clustering}.
We call it as Adaptive Neighborhood Feature Propagation (ANFP).
The optimal $S$ and $ F$ can be obtained via a simple algorithm which alternatively conducts the following {Step 1} and {Step 2} until convergence.

\noindent \textbf{Step 1}. Solve $S$ while fixing $F$, the problem becomes
\begin{align}
&\min_{S}\, \sum^n\nolimits_{i,j=1}D^x_{ij}S_{ij}+\gamma \|S\|^2_F+ \beta \mathrm{Tr} (F^{T}(I - S) F) \nonumber \\
& s.t. \ \ \sum^n\nolimits_{j=1}S_{ij}=1, S_{ij}\geq 0
\end{align}
which is equivalent to
\begin{align}
&\min_{S}\, \sum^n\nolimits_{i,j=1}(D^x-\beta FF^T)_{ij}S_{ij}+\gamma \|S\|^2_F \nonumber \\
& s.t. \ \ \sum^n\nolimits_{j=1}S_{ij}=1, S_{ij}\geq 0
\end{align}
This problem has a simple closed-form solution which is given as~\cite{Nie2014Clustering},
\begin{equation}
S_{ij} = \max\big\{-\frac{1}{2\gamma}(D^x - \beta FF^T)_{ij} + \eta, 0\big\}
\end{equation}
where $\eta = \frac{1}{k} +\frac{1}{2k\gamma}\sum^k\nolimits_{j=1} D^x_{ij}$.

\noindent \textbf{Remark.} Without loss of generality, here we suppose that $D^x_{i1}, D^x_{i2}\cdots D^x_{in}$ are ordered from small to large values, as discussed in work~\cite{Nie2014Clustering}.

\noindent \textbf{Step 2}. Solve $F$ while fixing $S$, the problem becomes
\begin{equation}
\min_{F} \mathrm{Tr} (F^{T}(I - S) F) + \mu \|F - H\|^2_F
\end{equation}
This is similar to Eq.(10) and the optimal solution is
%
\begin{equation}\label{EQ:C}
F = (1-\alpha)(I - \alpha S)^{-1} H
\end{equation}
%


\section{ANGPN Architecture}

In this section, we present our Adaptive Neighborhood Graph Propagation Network (ANGPN) based on the above ANFP model formulation.
Similar to the overall architecture of traditional GCN~\cite{kipf2016semi},
 we employ  one input layer, several hidden propagation layers and one final perceptron layer for ANGPN. They are introduced below.
%
%
%

\subsection{Hidden propagation layer}

Given an input feature matrix ${H}^{(k)}\in \mathbb{R}^{n\times d_k}$ and pair-wise distance matrix $D^x\in \mathbb{R}^{n\times n}$,
ANGPN hidden layer outputs feature map matrix ${H}^{(k+1)} \in \mathbb{R}^{n\times d_{k+1}}$ by using the above ANFP and non-linear transformation.
%
More specifically, let $F = \mathcal{P}(D^x, H)$ be the optimal solution of ANFP, then our ANGPN conducts layer-wise propagation in hidden layer as
\begin{align}\label{EQ:layer_gcn}
{H}^{(k+1)} = \sigma\big( \mathcal{P}(D^x,{H}^{(k)}){W}^{(k)}\big)
\end{align}
where ${D}^x$ denotes the distance matrix and ${W}^{(k)}\in \mathbb{R}^{d_{k}\times d_{k+1}}$ is the $k$-th layer-specific trainable weight matrix.
$\sigma(\cdot)$ denotes an activation function. 
%

%
 Directly calculating ANFP solution $F = \mathcal{P}(D^x, H)$ is time consuming due to (A) inversion operation in computing $F$ (Eq.(\ref{EQ:C})) and (B) alternative computation of {Step 1} and {Step 2} until convergence.
To reduce the issue of high computational complexity, similar to~\cite{jiang2019graph}, we derive an approximate algorithm to conduct them in the following.
For inversion operation (A), we adopt the strategy similar to GCN~\cite{kipf2016semi} and approximate the optimization of $F$ via an one-step power iteration algorithm. 
More concretely, instead of calculating Eq.(\ref{EQ:C}) directly, we use the following update to compute $F$.
\begin{equation}\label{EQ:propagation_rwr}
{F} = \alpha {S} {H} + (1-\alpha){H} = (\alpha S + (1-\alpha) I)H
\end{equation}
%
For computation {{(B)}}, we can also use a truncated $T$-step alternative iteration to optimize ANFP approximately.
%
Based on the above analysis, we summarize the whole propagation algorithm to compute $\mathcal{P}(D^x, H^{(k)})$ in hidden propagation layer of ANGPN in Algorithm 1.
%
%
%


%
\begin{algorithm}[h]
\caption{ANGPN propagation layer}
\begin{algorithmic}[1]
\STATE \textbf{Input:} Feature matrix ${H}^{(k)}\in \mathbb{R}^{n\times d_k}$ and ordered distances $D\in \mathbb{R}^{n\times n}$, parameters $\gamma, \beta$ and $\alpha$,  maximum iteration $T$
\STATE \textbf{Output:} Feature map $H^{(k+1)}$
\STATE Initialize $F = H^{(k)}$
\STATE Compute $\eta$ as 
$
\textstyle \eta = \frac{1}{k} +\frac{1}{2k\gamma}\sum^k\nolimits_{j=1} D^x_{ij}
$
\FOR {$t=1,2\cdots T$}
\STATE Compute $S$ as\\
$
\ \ \ \ \ \ \ \ \ \ S_{ij} = \max\big\{-\frac{1}{2\gamma}(D^x - \beta FF^T)_{ij} + \eta, 0\big\}
$
\STATE Compute $F$ as\\
$
\ \ \ \ \ \ \ \ \ \ F = (\alpha S + (1-\alpha) I)H^{(k)}
$
\ENDFOR \label{code:recentEnd}
\STATE Return
$
\mathcal{P}(D^x, H^{(k)}) ={F}
$
\emph{}
\end{algorithmic}
\end{algorithm}


\subsection{Perceptron layer}

For graph node classification, the last layer of ANGPN outputs the final label prediction $Z\in \mathbb{R}^{n\times c}$ for graph nodes where $c$ denotes the number of class.
We add a softmax activation function 
  on each row of the final output feature map  ${H}^{(K)} \in \mathbb{R}^{n\times c}$ as,
 \begin{equation}\label{EQ:softmax}
 {Z}=\textrm{softmax}({H}^{(K)})
 \end{equation}
where each row ${Z}_i$ of matrix ${Z}$  denotes the corresponding label prediction vector for the $i$-th node.
%
%
The optimal network weight parameters $\{{W}^{(0)},{W}^{(1)},\cdots {W}^{(K-1)}\}$ of ANGPN are trained  by minimizing  cross-entropy loss function defined as~\cite{kipf2016semi},
 \begin{equation}
\mathcal{L}_{\textrm{Semi-ANGPN}} = -\sum\nolimits_{i\in L} \sum^c\nolimits_{j=1} {Y}_{ij}\mathrm{ln} {Z}_{ij}
 \end{equation}
where ${L}$ indicates the set of labeled nodes and ${Y}_{i\cdot}, i\in L$ denotes the corresponding label indication vector for the $i$-th labeled node. 
%

\subsection{Computational complexity}

Empirically, the optimization in each layer of ANGPN generally does not lead to very high computational cost because it can be solved via a simple (efficient) algorithm (Algorithm 1) in practical. Overall, ANGPN has similar time consuming  with GAT~\cite{velickovic2017graph}. However, ANGPN outperforms GAT (shown in Table I).
Theoretically, in each iteration, the complexity of our graph construction and convolution are $O(kn)+O(dn^2)$ and $O(n^2 d)$, where $n,d$ denote the graph size and feature dimension in each layer. The overall complexity is $O(T (k n+2d n^2))$, where $T$ is number of maximum steps. In our experiments, we set $T=2$ which can return desired better results.  

\section{Experiments}


\subsection{Datasets}

We test our method on four benchmark datasets, including SVHN~\cite{netzer2011reading}, 20News~\cite{Lang95}, CIFAR~\cite{krizhevsky2009learning} and CoraML~\cite{sen2008collective}.
The details of these datasets and their usages in our experiments are introduced below. 

\noindent\textbf{SVHN.}
This dataset contains 73257 training and 26032 testing images~\cite{netzer2011reading}.
Each image is a $32\times 32$ RGB color image which contains  multiple number of  digits and
the task is to recognize the digit in the image center.
In our experiments, we select 500 images for each class and
obtain 5000 images in all for our evaluation.
We have not use all of images because it requires large storage and high computational complexity for graph convolution operation in our ANGPN and other related GCN based methods.
We extract a commonly used CNN feature descriptor for each image data.

\noindent\textbf{CIFAR.}
This dataset contains  50000 natural images which are falling into 10 classes~\cite{krizhevsky2009learning}.
Each image in this dataset is a $32\times 32$ RGB color image.
In our experiments, we select 500 images for each class and use 5000 images in all for
our evaluation. For each image, we extract a widely used CNN feature descriptor for it.

\noindent\textbf{20News.}
The 20 Newsgroups data set is a collection of approximately 20,000 newsgroup documents, which are
partitioned nearly across 20 different newsgroups~\cite{Lang95}.
In our experiments, we use a subset with 3970 data points in 4 classes.
Each data is represented by a 8014 dimension feature descriptor.

\noindent\textbf{CoraML.}
The CoraML dataset consists of 1617 scientific publication documents which are classified into one of seven class.
Each document is represented by a 8329 dimension feature descriptor.
\\

\begin{table*}[!htp]
\centering
\caption{Comparison results of different methods on four benchmark datasets. The best results are marked as bold. } \label{b1}
\centering
\begin{tabular}{c||c|c|c||c|c|c}
  \hline
  \hline
  Dataset& \multicolumn{ 3}{c||}{SVHN} & \multicolumn{ 3}{c}{CIFAR}\\
  \hline
  No. of label & 10\% & 20\% & 30\% & 10\% & 20\% & 30\%  \\
  \hline
  ManiReg~\cite{belkin2006manifold} & 71.91$\pm$ 1.50 & 76.43 $\pm$ 0.63 & 78.67 $\pm$ 0.71 & 59.44 $\pm$ 1.24 & 65.11 $\pm$ 0.79 &\textbf{66.04 $\pm$ 0.58} \\
  LP~\cite{zhu2003semi} & 55.56 $\pm$ 0.62 & 66.29 $\pm$ 0.78 & 70.52 $\pm$ 0.72 & 52.66 $\pm$ 1.40 & 57.66 $\pm$ 0.39 & 60.36 $\pm$ 1.05 \\
  DeepWalk~\cite{perozzi2014deepwalk} & 74.54 $\pm$ 0.87 & 76.72 $\pm$ 0.83 & 77.35 $\pm$ 1.36 & 58.82 $\pm$ 0.96 & 61.62 $\pm$ 0.54 & 63.76 $\pm$ 0.82 \\
  DGI~\cite{velickovic2018deep} & 72.05 $\pm$ 0.99 & 74.83 $\pm$ 0.46 & 75.77 $\pm$ 0.64 & 58.18 $\pm$ 0.57 & 61.66 $\pm$ 0.18 & 64.08 $\pm$ 0.39  \\
  GraphSage~\cite{hamilton2017inductive} & 73.70 $\pm$ 0.92 & 77.22 $\pm$ 0.58 & 78.78 $\pm$ 0.31 & 61.85 $\pm$ 0.98 & 63.58 $\pm$ 0.68 & 65.21 $\pm$ 0.38 \\
  GCN~\cite{kipf2016semi} & 75.10 $\pm$ 0.60 & 76.45 $\pm$ 0.61 & 77.27 $\pm$ 0.57 & 60.36 $\pm$ 1.00 & 63.17 $\pm$ 0.61 & 63.87 $\pm$ 0.56  \\
  GAT~\cite{velickovic2017graph} & 72.60 $\pm$ 0.81 & 75.13 $\pm$ 0.38 &75.17 $\pm$ 0.59 & 59.62 $\pm$ 0.35 & 61.09 $\pm$ 0.54 & 62.19 $\pm$ 0.34 \\
  \hline
  ANGPN & \textbf{76.85 $\pm$ 0.83} & \textbf{78.69 $\pm$ 0.53} & \textbf{79.62 $\pm$ 0.55} & \textbf{62.42 $\pm$ 1.04} & \textbf{65.43 $\pm$ 0.66} &65.90 $\pm$ 0.49  \\
  \hline
  \hline
  Dataset & \multicolumn{ 3}{c||}{20News} & \multicolumn{ 3}{c}{CoraML}\\
  \hline
  No. of label  & 10\% & 20\% & 30\% & 10\% & 20\% & 30\%  \\
  \hline
  ManiReg~\cite{belkin2006manifold}  & 89.80 $\pm$ 0.93 & 91.08 $\pm$ 0.41 & 91.85 $\pm$ 0.23 & 44.90 $\pm$ 1.85 & 58.00 $\pm$ 1.64 & 64.61 $\pm$ 1.60 \\
  LP~\cite{zhu2003semi} & 88.34 $\pm$ 1.35 & 90.92 $\pm$ 0.35 & 91.55 $\pm$ 0.81 & 54.38 $\pm$ 1.93 & 61.42 $\pm$ 1.15 & 64.42 $\pm$ 0.84 \\
  DeepWalk~\cite{perozzi2014deepwalk}  & 88.62 $\pm$ 0.70 & 90.31 $\pm$ 0.66 & 91.12 $\pm$ 0.38 & 60.28 $\pm$ 2.59 & 64.14 $\pm$ 1.63 & 66.17 $\pm$ 1.76 \\
  DGI~\cite{velickovic2018deep} & 89.47 $\pm$ 0.69 & 90.76 $\pm$ 0.90 & 91.02 $\pm$ 0.36 & 62.88 $\pm$ 0.86 & 65.16 $\pm$ 0.60 & 68.43 $\pm$ 1.01  \\
  GraphSage~\cite{hamilton2017inductive} & 90.46 $\pm$ 0.55 & 91.79 $\pm$ 0.73 & 92.13 $\pm$ 0.44 & 63.51 $\pm$ 2.01 & 68.43 $\pm$ 0.46 & 68.91 $\pm$ 0.65 \\
  GCN~\cite{kipf2016semi} & 89.94 $\pm$ 0.49 & 91.40 $\pm$ 0.46 & 91.45 $\pm$ 0.74 & 63.33 $\pm$ 1.44 & 68.22 $\pm$ 0.72 & 68.16 $\pm$ 1.18  \\
  GAT~\cite{velickovic2017graph}  & 88.00 $\pm$ 0.16 & 89.07 $\pm$ 0.47 & 89.05 $\pm$ 1.10 & 58.16 $\pm$ 1.76 & 60.77 $\pm$ 0.55 & 61.64 $\pm$ 1.53 \\
  \hline
  ANGPN & \textbf{90.80 $\pm$ 0.34} & \textbf{92.35 $\pm$ 0.37} & \textbf{92.83 $\pm$ 0.48} & \textbf{65.53 $\pm$ 1.25} & \textbf{69.22 $\pm$ 1.50} & \textbf{69.55 $\pm$ 1.44}  \\
  \hline
  \hline
\end{tabular}
\end{table*}

\subsection{Experimental setting}

%

For all datasets, we randomly select 10\%, 20\% and 30\% samples in each class as labeled  data for training the network and
use the other 5\% of labeled data for validation purpose.
 The remaining samples are used as the unlabeled test samples.
All the reported results are averaged over five runs  with different groups of training, validation and testing data splits.
The number of hidden convolution layers in  ANGPN is set as 2.
The number of units in each hidden layer is set as 50.
Some additional experiments across different number of convolution layers are presented in Parameter analysis.
We train our ANGPN for a maximum of 10000 epochs (training
iterations) by using an ADAM algorithm~\cite{Adam} initialized by Glorot initialization~\cite{glorot2010understanding} with learning rate 0.005.
The parameters $\{\beta,\alpha\}$ in ANGPN are set to $\{0.3,0.5\}$, respectively. 

\subsection{Comparison results}

\textbf{Baselines.} 
We  compare our method against some other  related graph approaches which contain i)  graph based
semi-supervised learning method Label Propagation (LP)~\cite{zhu2003semi}, Manifold Regularization (ManiReg)~\cite{belkin2006manifold},
and ii) graph neural network methods including DeepWalk ~\cite{perozzi2014deepwalk}, Graph Convolutional Network (GCN)~\cite{kipf2016semi},
Graph Attention Networks (GAT)~\cite{velickovic2017graph}, Deep Graph Informax(DGI)~\cite{velickovic2018deep} and GraphSAGE~\cite{hamilton2017inductive}.
The codes of these comparison methods were provided by authors and we use them directly on our data setting in experiments.
In addition, as discussed before,
the core aspect of the proposed ANGPN is that it conducts graph construction and graph convolution cooperatively to boost their respective performance.
To demonstrate the effectiveness of this cooperative learning,
we construct a baseline method, NGPN, which first constructs the graph $S$ via Eq.(15) with $\beta=0$ and then conducts feature propagation on $S$ via Eq.(19).

%
\textbf{Comparison results.} Table~\ref{b1} summarizes the comparison results on four benchmark datasets. 
The best results are marked as bold in Table~\ref{b1}.
%
We can note that,
(1) ANGPN outperforms the baseline method GCN~\cite{kipf2016semi} on all datasets, demonstrating the desired ability of the proposed ANGPN network on conducting semi-supervised learning tasks by learning neighborhood graph adaptively in data representation and learning.
(2) ANGPN consistently performs better than recent graph network model GAT ~\cite{velickovic2017graph}, Deep Graph Informax(DGI)~\cite{velickovic2018deep} and GraphSAGE~\cite{hamilton2017inductive}. 
(3) ANGPN can obtain better performance than other graph based semi-supervised learning methods, such as LP~\cite{zhu2003semi}, ManiReg~\cite{belkin2006manifold} and DeepWalk~\cite{perozzi2014deepwalk}. It demonstrates the effectiveness of ANGPN on conducting semi-supervised learning tasks.

Table~\ref{b2} shows the results of ANGPN vs. NGPN on four datasets.
Here, we can note that, ANGPN consistently performs better than
the baseline model NGPN.
This further demonstrates the advantages of ANGPN by conducting graph construction and graph convolution cooperatively and thus can boost their respective performance.



\begin{table}[!htp]\small
\centering
\caption{Comparison results of ANGPN vs. NGPN on four different benchmark datasets. }\label{b2}
\centering
\begin{tabular}{c|c|c|c|c}
\hline
\hline
  Dataset &method& 10\% & 20\% & 30\%  \\
 \hline
  \multirow{2}{*}{SVHN}&NGPN&75.73$\pm$0.67&77.72$\pm$0.64&78.49$\pm$0.59 \\
  \cline{2-5}
  &ANGPN&76.85$\pm$0.83&78.69$\pm$0.53&79.62$\pm$0.55 \\
 \hline

 \multirow{2}{*}{CIFAR} &NGPN&61.41$\pm$0.77&64.51$\pm$0.74&64.65$\pm$0.58 \\
 \cline{2-5}
  &ANGPN&62.42$\pm$1.04&65.43$\pm$0.66&65.90$\pm$0.49 \\
 \hline

  \multirow{2}{*}{20News} &NGPN&90.01$\pm$0.35 &91.64$\pm$0.54&91.97$\pm$0.71 \\
 \cline{2-5}
  &ANGPN&90.80$\pm$0.34&92.35$\pm$0.37&92.83$\pm$0.48 \\
 \hline

   \multirow{2}{*}{CoraML}&NGPN&63.42$\pm$1.75&66.40$\pm$0.71&67.88$\pm$1.92 \\
 \cline{2-5}
 &ANGPN&65.53$\pm$1.25&69.22$\pm$1.50&69.55$\pm$1.44 \\
  \hline
  \hline
\end{tabular}
\end{table}
\begin{figure}[!htpb]
\centering
\centering
\includegraphics[width=0.24\textwidth]{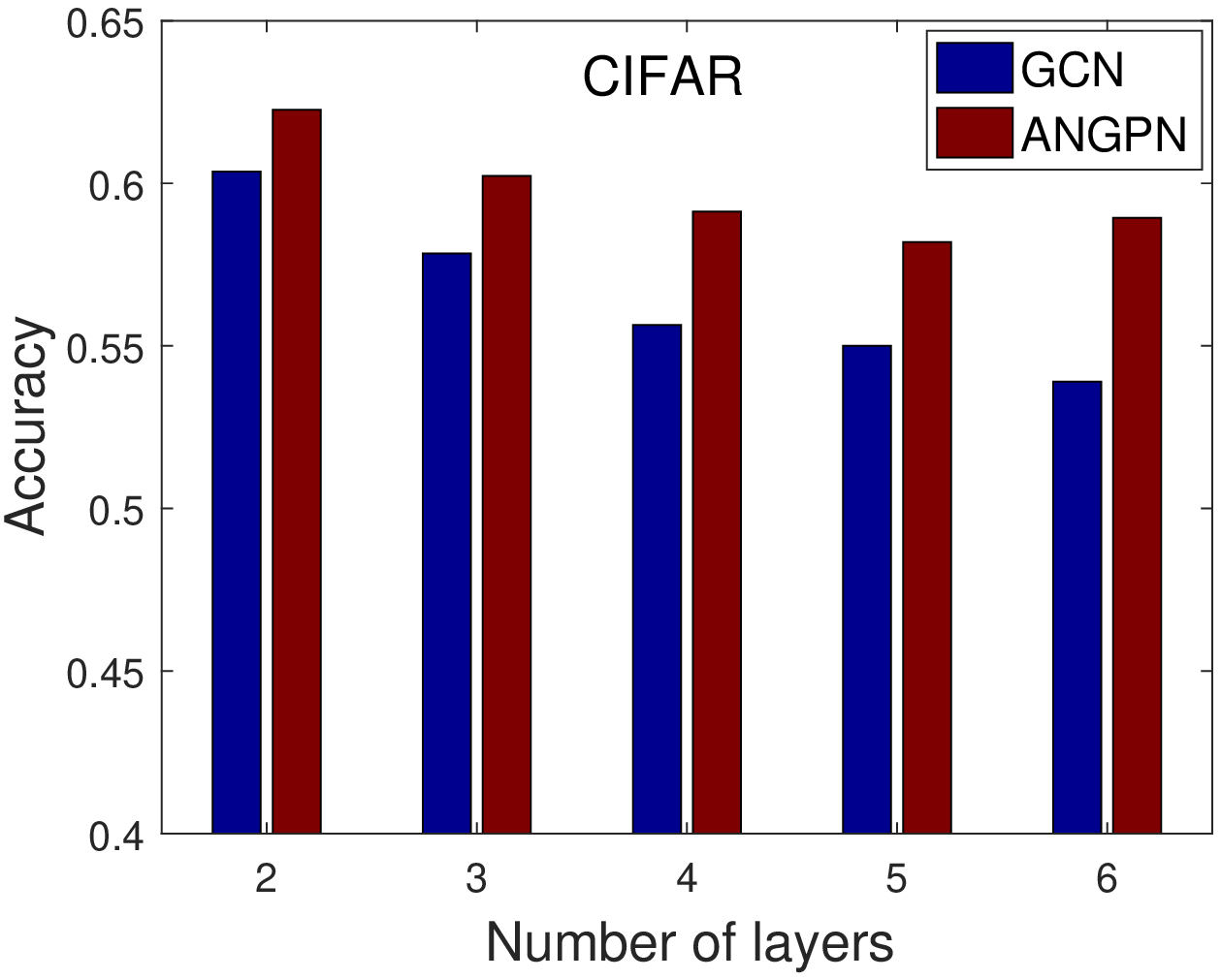}\includegraphics[width=0.24\textwidth]{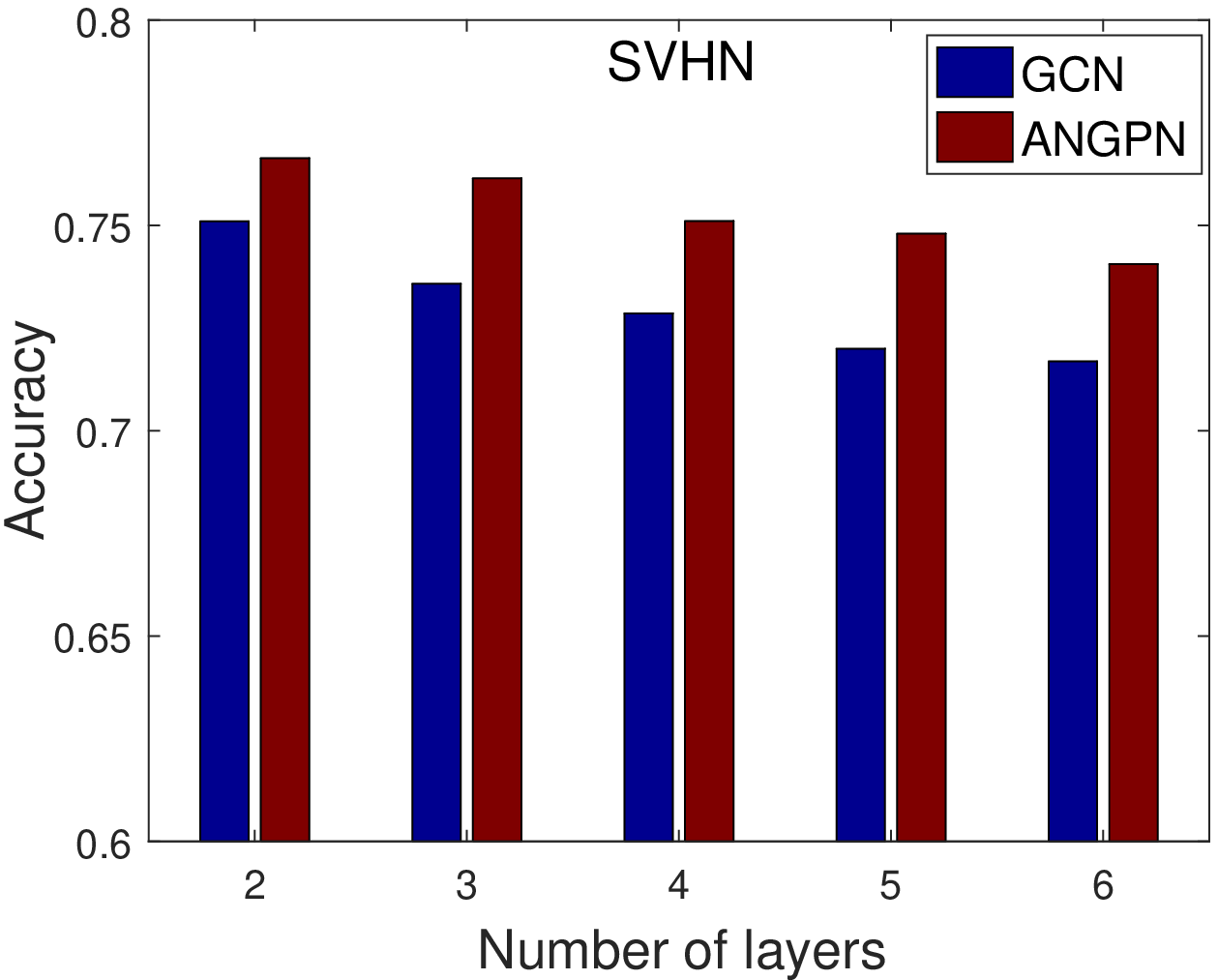}
\includegraphics[width=0.24\textwidth]{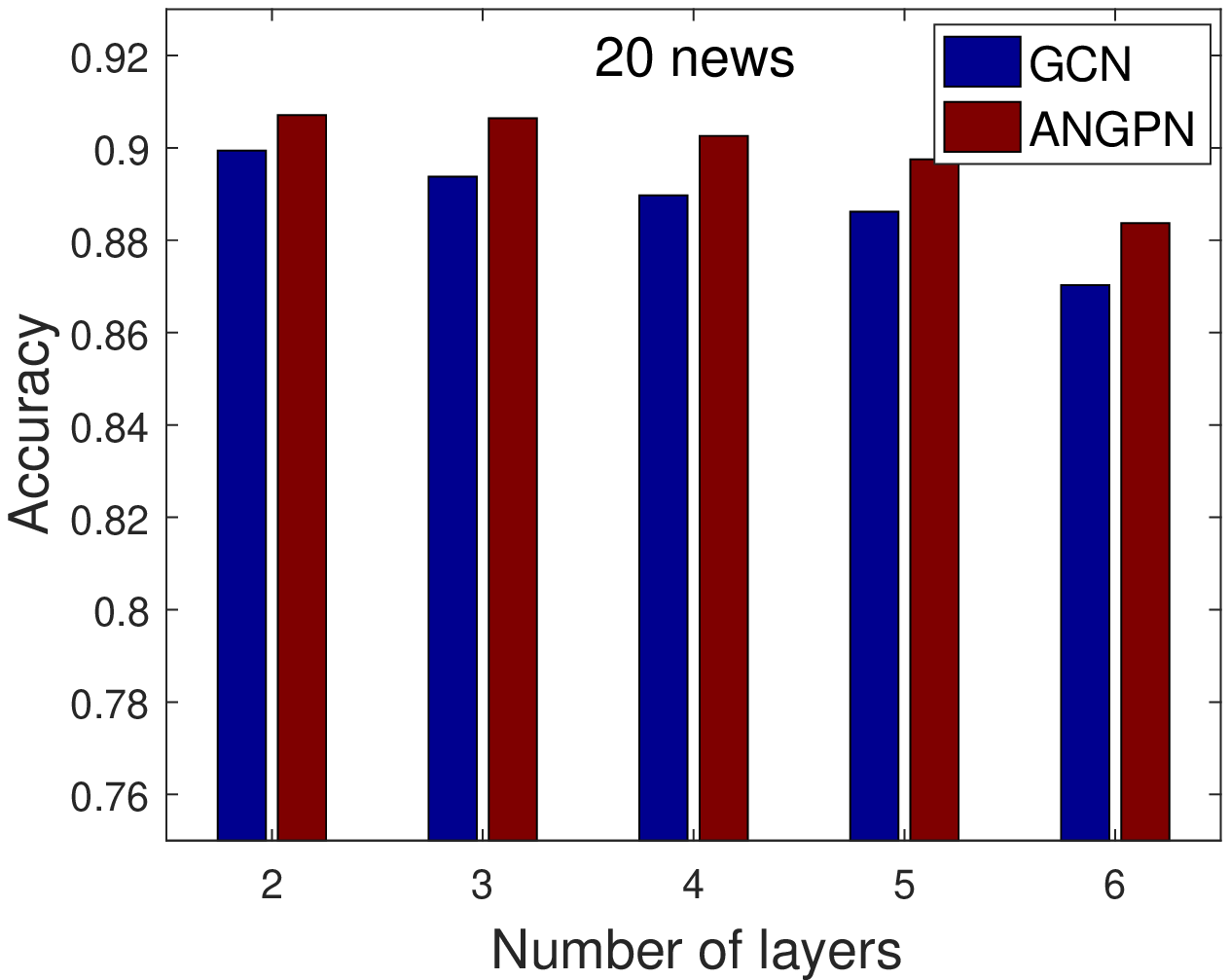}\includegraphics[width=0.24\textwidth]{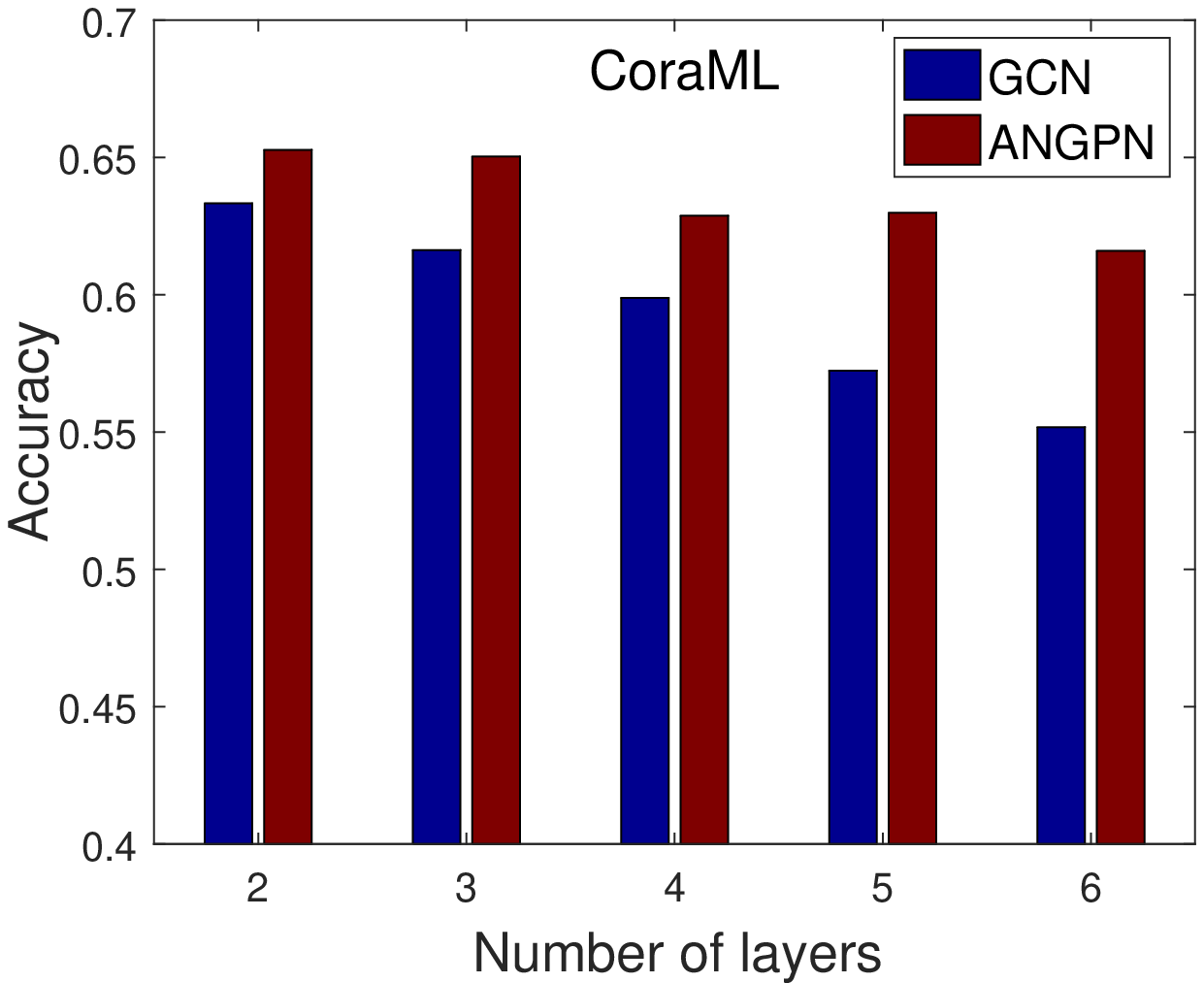}
  \caption{Results of ANGPN across different number of convolutional layers on four datasets.}\label{fig::lambda}
\end{figure}
%

%

\subsection{Parameter analysis}

In this section, we evaluate the performance of ANGPN model with different network settings.  
%
Figure 2 shows the performance of ANGPN method across different number of convolutional layers on four datasets, respectively.
One can note that
ANGPN can obtain better performance with different number of layers, which indicates the insensitivity of ANGPN w.r.t. model depth.
Also, ANGPN always performs better than GCN under different model depths, which further
demonstrates the benefit and better performance of ANGPN comparing with the baseline method.

\section{Conclusion}

This paper proposes a novel Adaptive Neighborhood Graph Propagation Network (ANGPN) for graph based semi-supervised learning problem.
ANGPN integrates neighborhood graph construction and graph convolution architecture together in a unified formulation, which
can learn an optimal neighborhood graph structure that best serves the proposed graph propagation network for data representation and semi-supervised learning problem.
Experimental results on four benchmark datasets demonstrate the effectiveness and advantage of ANGPN model on various semi-supervised learning tasks.
%
In the future,, we will explore ANGPN model for some other machine learning tasks, such as graph embedding, data clustering etc. 


\bibliographystyle{IEEEtran}
\bibliography{nmfgm}

\end{document}